\documentclass[12pt,letter]{article}

\usepackage{amsmath,amsfonts,bm}

\def\eqref#1{equation~\ref{#1}}

\def\1{\bm{1}}

\def\rvh{{\mathbf{h}}}

\def\rvx{{\mathbf{x}}}
\def\rvy{{\mathbf{y}}}

\DeclareMathAlphabet{\mathsfit}{\encodingdefault}{\sfdefault}{m}{sl}
\SetMathAlphabet{\mathsfit}{bold}{\encodingdefault}{\sfdefault}{bx}{n}

\usepackage{microtype}
\usepackage{graphicx}
\usepackage{subfigure}
\usepackage{booktabs}       
\usepackage{amsfonts}       
\usepackage{nicefrac}       
\usepackage{algorithm}
\usepackage{algorithmic}
\usepackage{amsmath}
\usepackage{amssymb}

\usepackage[dvips]{epsfig}
\usepackage{epstopdf}
\usepackage{enumitem}
\usepackage{amsthm}
\usepackage{array}
\usepackage{color}

\usepackage[paperwidth=8.5in,left=1.0in,right=1.0in,top=1.0in,bottom=1.0in,
paperheight=11.0in]{geometry}

\usepackage{hyperref}
\usepackage{url}

\title{Mixed Membership Recurrent Neural Networks}

\author{Ghazal Fazelnia
	\thanks{Department of Electrical Engineering, Columbia University. Email: {\ttfamily ghazal@ee.columbia.edu}}\\
	\and Mark Ibrahim
	\thanks{Center for Machine Learning, Capital One. Email: {\ttfamily mark.ibrahim@capitalone.com}}\\
	\and Ceena Modarres
	\thanks{Center for Machine Learning, Capital One. Email: {\ttfamily ceena.modarres@capitalone.com}}\\
	\and Kevin Wu
	\thanks{Center for Machine Learning, Capital One. Email: {\ttfamily kevinxwu@gmail.com}}\\
	\and John Paisley\thanks{Department of Electrical Engineering, Columbia University. Email: {\ttfamily jpaisley@columbia.edu}}
}

\date{}

\begin{document}
\maketitle

\begin{abstract}
Models for sequential data such as the recurrent neural network (RNN) often implicitly model a sequence as having a fixed time interval between observations and do not account for group-level effects when multiple sequences are observed. We propose a model for grouped sequential data based on the RNN that accounts for varying time intervals between observations in a sequence by learning a group-level base parameter to which each sequence can revert. Our approach is motivated by the mixed membership framework, and we show how it can be used for dynamic topic modeling in which the distribution on topics (not the topics themselves) are evolving in time. We demonstrate our approach on a dataset of 3.4 million online grocery shopping orders made by 206K customers. 
\end{abstract}

\section{Introduction}
Recurrent neural networks (RNNs) have become one of the standard models in sequential data analysis \cite{rumelhart1986learning,elman1990finding}. At each time step of the RNN, an observation is modeled via a neural network using the observations and hidden states from previous time points. Models such as the RNN, and also the hidden Markov model among others, often implicitly assume a sequence as having a fixed time interval between observations. They also often do not account for group-level effects when multiple sequences are observed and each sequence belongs to one of multiple groups.

For example, consider data in the form of a sequence of discrete counts by a set of groups---e.g., a sequence of purchases (market baskets) for a set of customers, with one sequence per customer. A vanilla RNN implementation would model these sequences using a network with the same parameters, which removes the customer-level information, and according to an enumerated indexing, which removes the time interval information between orders. However, this information is important: customer-specific effects can improve predictive performance for each customer, while an interval of one day versus one month between orders significantly impacts the items likely to be purchased next.

While a few methods have been been proposed, there is no standard technique for addressing these shortcomings. Most previous work has focused on filling in the missing values, such as with zeros or the global mean of the data, also known as data imputation \cite{bengio1996recurrent,tresp1998solution,parveen2002speech}. Recent work proposed in \cite{lipton2016directly,choi2016doctor} also impute with either zeros or the previous observed value. In \cite{che2018recurrent} the authors propose a method to incorporate the missingness pattern via a decay parameter and masking technique with a modified gated recurrent unit (GRU), but this assumes a single data sequence and is not easily modified to learn local effects for groups of sequences.

Our perspective is to assume that, as more time passes between any two observations, the value of the sequential information for making the next prediction decreases. For example, in online shopping, orders made in consecutive days may have better sequential predictive power than orders made month by month. If a customer takes a 6-month shopping break, the previous sequence from that customer may be valueless in predicting the next order, and instead a customer-specific initial distribution may be more useful for predicting what is the start of a new sequence.

In this paper we propose a sequential modeling approach that can be viewed as a continuous-time mixed membership RNN. In this approach to modeling multiple groups of sequential data, each group shares the same RNN parameters while having a group-specific bias to which that group can revert when starting a new sequence. We use a time-dependent weighting to smoothly transition from this globally-shared sequential model to the group-level starting model that considers how much time has elapsed between each observation.

We are motivated by text models such as latent Dirichlet allocation (LDA), which learns a global set of semantically meaningful topics and local distributions for each group \cite{blei2003latent}. We discuss two variations on our approach, one that directly models the sequence of observations and one that models the sequence of topic distributions. The second lets us view our method as a dynamic topic model, which differs from that proposed by \cite{blei2006dynamic} by modeling sequences of topic distributions over a fixed set of topics, rather than evolving topics for use at different time points by new observations.

The rest of the paper is organized as follows: In Section \ref{sec.background} we review basic RNN and mixed membership modeling background. In Section \ref{sec.mm_rnn} we present our proposed mixed membership RNN (MM-RNN) approach, one basic data-level modeling approach and one extension to the dynamic topic modeling problem. In Section \ref{sec.experiments} we show experiments on the Instacart online shopping data set.

\section{Background}\label{sec.background}
\subsection{Recurrent Neural Networks}
A recurrent neural network models a sequence of vectors $\rvy = (y_1,\dots,y_T)$ with a neural network that takes as input a corresponding sequence of vectors $\rvx = (x_1,\dots,x_T)$ along with internal hidden states from the network, $\rvh = (h_1,\dots,h_T)$. When the model is probabilistic, this can be viewed as defining a joint likelihood of the data,
\begin{equation}\label{eq.1}
 p(\rvy|\theta) = p(y_1|\theta) \prod_{t=2}^{T} p(y_{t}|y_{1:t-1},\theta),
\end{equation}
where $p(y_{t}|y_{1:t-1},\theta) \equiv p(y_t|h_t)$ and 
\begin{equation}
h_t = f_{\theta}(x_t,h_{t-1})~~{\text{for}}~t > 0,
\end{equation}
usually with $x_t \equiv y_{t-1}$. The non-linear function $f_{\theta}$ can be a standard RNN cell, or a more complex GRU \cite{cho2014learning} or LSTM \cite{hochreiter1997long}, and $\theta$ are its parameters. We will use a function of the form
$$f_{\theta}(x_t,h_{t-1}) \equiv f_{\hat{\theta}}(Wx_t + Uh_{t-1}),$$
where $W$ and $U$ are matrices. Possible forms of the distribution $p(y_t|h_t)$ include Gaussian, multinomial and Poisson, as determined by the problem. The goal is typically to do maximum likelihood or MAP inference over Eq.\ (\ref{eq.1}), depending on whether priors are on $\theta$.

When multiple sequences are observed, a typical and straightforward approach is to treat them as independent from the same RNN,
\begin{equation}
    p(\rvy_1,\dots,\rvy_D|\theta) = \prod_{d=1}^D  p(\rvy_d|\theta).
\end{equation}
To account for variations in sequences across multiple groups (for example, indexed by $D$ above), mixtures of RNNs are one straightforward approach. In this paper we take a different approach motivated by the mixed membership modeling framework described below.

\subsection{Mixed membership models}
Mixed membership models provide a probabilistic approach to modeling groups of data through a combination of shared and group-specific parameters \cite{airoldi2014handbook}. The best known mixed membership model is latent Dirichlet allocation (LDA) \cite{blei2003latent}, but many variations exist. 

The basic structure of a mixed membership model is:
\begin{enumerate}
    \item Generate global variables $\theta \sim p(\theta)$.
    \item For the $d$th group of data:
    \begin{enumerate}
        \item Generate local variables $\phi_d \sim p(\phi)$
        \item Generate data $y_d \sim p(y_d|\phi_d,\theta)$
    \end{enumerate}
\end{enumerate}
The distribution $p(y_d|\phi_d,\theta)$ is a mixture where the variables $\theta$ define the globally shared set of distributions and $\phi_d$ is used to define the weights on these distributions.

LDA and related models let $\theta = \{\beta_1,\dots,\beta_k\}$ be a set of distributions on a discrete item set (often words in a vocabulary), and $\phi_d$ be a probability vector on those topics. One example closely related to our work is the correlated topic model (CTM) \cite{blei2007correlated}, in which $\phi_d \sim \mathcal{N}(\mu,\Sigma)$, and a softmax function $\sigma(\phi_d)$ transforms this vector into a distribution on topics. 

A key benefit of such models is that each group of data can mix over the same set of distributions, allowing them to share statistical strength during inference, while also allowing a meaningful comparison across groups via their shared representation in these distributions. In the next section, we are motivated by this mixed membership modeling framework when defining a shared RNN for multiple sequences that also allows each sequence to have its own unique characteristics.

\begin{figure*}[t!]
    \centering
    \includegraphics[width=.95\textwidth]{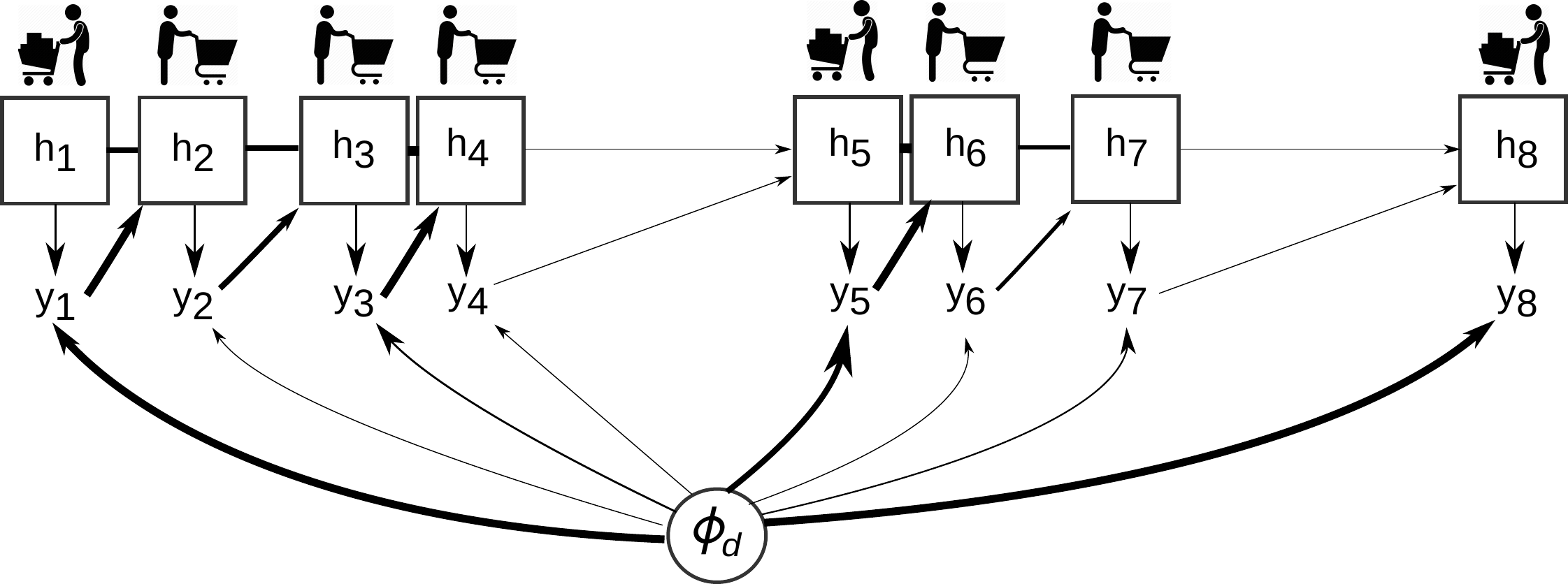}
    \caption{The unrolled proposed framework we use in our experiments, shown for the $d$th sequence. Each prediction uses a weighted combination of an RNN and a customer-specific parameter. As the time between observations decreases the RNN prediction is favored more heavily, while as the time lag increases the prediction is more based on an independent draw from a distribution using the parameter $\phi_d$ learned for person $d$. Those effects are shown by thicker arrows in this figure.}
    \label{fig:my_label}
\end{figure*}

\section{Mixed Membership RNN Models}\label{sec.mm_rnn}
\subsection{Motivation}
Recurrent neural networks have shown state-of-the-art performance for modeling sequential data, but have difficulty in capturing global semantic information. On the other hand, topic models have the ability to capture global semantics, but are usually not sequential models and lack the modeling power of the RNN in this regard. Recent work by \cite{dieng2016topicrnn} has demonstrated the advantage of combining these two modeling paradigms for natural language models of text. Inspired by this perspective, we address a significantly different problem.

One challenge faced by the basic RNN framework comes when the data is temporal and the time step between consecutive values is not equal---an issue avoided in natural language models of text. For example, a sequence of grocery orders by a customer will likely not be equally spaced in time, and it is obvious that as more time passes since the last order, the distribution on items in the next order should change. A vanilla implementation of the RNN on such data would ignore the time component and index orders by number. If the data is coarsely aggregated, for example by day, one solution might be to treat this as a missing data problem. For many problems a ``missing'' observation can also be argued to be an observation of all zeros, which can be directly used as data for the RNN. 

Instead, we are motivated by problems in which a long delay in time between observations indicates that the sequential information in the previous observations loses its value for predicting the next observation. But because we are interested in grouped data structures, the model should not revert to the same base prediction but should be group specific. We have discussed the shopping scenario with multiple customers, but this appears in other settings as well; a very long delay in time between audio speech by a user (e.g., to Siri, Alexa, etc.) would indicate that the previous sequence has ended, but the distribution on the start of the next sequence should vary based on the user.

To this end, we propose a model that accounts for the following: 1) each prediction is influenced by both the previous sequential information as well as biases that are group-specific; 2) as the time intervals increase, the prediction smoothly adapts by tending to be close to this default bias prediction and further from what the purely sequential prediction would be. For example, in a dynamic topic modeling problem in which the topics are fixed and a sequence of topic distributions are generated, the topic distribution smoothly reverts to a group-specific base topic distribution as the time between documents increases.

\subsection{The basic framework}
\label{sec.basic_model}

We first present the basic idea of the model directly on data $\rvy_d$, $d=1,\dots,D$, where each $\rvy_d$ is a sequence of vectors with corresponding sequence of time stamps. In this model, we define $\rho(\Delta t) \in [0,1]$ to be a function of the time interval between two particular observations in a sequence, $\Delta t$. This value produces a weighted average and decreases as $\Delta t$ increases. For example, in our experiments we use $\rho(\Delta t) = (t_0 + \Delta t)^{-\kappa}$ with $t_0,\kappa > 0$. $\rho(\cdot)$ will allow us to define a continuous-time RNN that adjusts to periods of no observations.

\begin{algorithm}\vspace{5pt}
\textbf{Basic MM-RNN:} Let $\rho(\Delta t) \in [0,1]$ be decreasing in $\Delta t$. For chosen RNN cell $f_{\theta}$ and arbitrary function $\sigma$,\vspace{-5pt}
\begin{enumerate}
    \item Generate RNN parameters $\theta\sim p(\theta)$\vspace{-5pt}
    \item Generate group-level bias vectors $\phi_d \sim_{iid} \mathcal{N}(\mu,\Sigma)$\vspace{-5pt}
    \item For $d$th group, generate the sequence $\rvy_d$. Dropping several $d$ indexes, at step $t$ in this sequence,\vspace{-2pt}
    \begin{enumerate}
        \item Compute $h_t = f_{\theta}(x_t,h_{t-1})$, e.g., $x_t \equiv y_{t-1}$
        \item Compute $\sigma_t \equiv \sigma(\rho_t h_t + (1-\rho_t)\phi_d)$
        \item Generate $y_t \sim p(y|\sigma_t)$
    \end{enumerate}
\end{enumerate}
\end{algorithm}

\begin{figure}[b]
    \centering
    \includegraphics[width=.5\columnwidth]{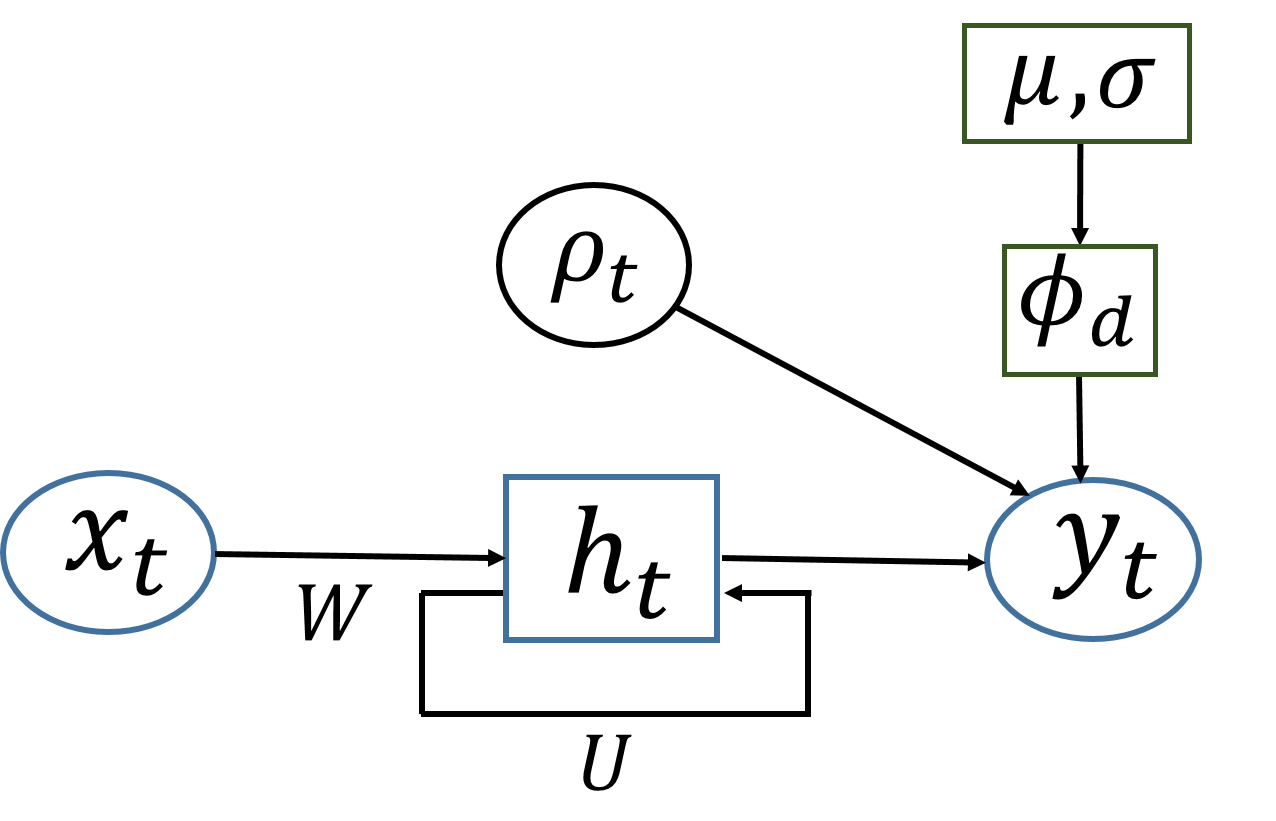}
    \caption{Graphical model of the basic network.}
    \label{fig:network-graph}
\end{figure}

The basic MM-RNN model is shown in the table above. To give two specific examples, if $y_t$ were a histogram of counts, then $\sigma_t$ could be the softmax function and $p(y|\sigma)$ a multinomial leading to the cross entropy penalty. Or $p(y|\sigma)$ could be a (technically inappropriate) Gaussian distribution on the normalized $y$ with $\sigma$ as the mean, resulting in an L2 penalty.

In the proposed framework, we modify the RNN by including a group-specific bias vector $\phi_d \in \mathbb{R}^K$. Then, rather than generate $\rvy_d(t)$ dependent on $\rvh_d(t)$ as in the typical RNN setup, in Step 3(b) we average $\rvh_d(t)$ with $\phi_d$ according to the function $\rho$. As discussed, $\rho$ decreases as the time interval between $\rvy_d(t-1)$ and $\rvy_d(t)$ increases. When $\rho=0$, $\rvy_d(t)$ is independently generated from the base distribution for group $d$. The definition of $\rho(\cdot)$ determines the rate at which the RNN \textit{is forgotten}; the RNN can have its own forgetting mechanism as well. When $\rho = 1$ the sequence is being fully modeled by an RNN. We show the basic graphical model of our network in Figure \ref{fig:network-graph}.

We anticipate that this approach can give better predictions by: 1) not artificially learning sequential information that it isn't there, and 2) allowing a better RNN to be learned by focusing on the part of the data where sequential information is present, which we consider to be when the time between observations is short.

\begin{algorithm}[t]\vspace{5pt}
\textbf{MM-RNN topic model:} Let $\rho(\Delta t) \in [0,1]$ be decreasing in $\Delta t$. For a chosen RNN cell $f_{\theta}$,\vspace{-5pt}
\begin{enumerate}
    \item Generate RNN parameters $\theta\sim p(\theta)$\vspace{-5pt}
    \item Generate group-level bias vectors $\phi_d \sim_{iid}    \mathcal{N}(\mu,\Sigma)$\vspace{-5pt}
    \item Generate topics $\beta_k \sim \mathrm{Dirichlet}(\alpha),~ k=1,\dots,K$\vspace{-5pt}
    \item For $d$th group, generate the sequence $\rvy_d$. Dropping several $d$ indexes, at step $t$ in this sequence,\vspace{-2pt}
    \begin{enumerate}
        \item Compute $h_t = f_{\theta}(x_t,h_{t-1})$, e.g., $x_t \equiv y_{t-1}$
        \item Compute $\sigma_t = \mathrm{softmax}(\rho_t h_t + (1-\rho_t)\phi_d)$
        \item Generate $y_t \sim p(y|\boldsymbol{\beta},\sigma_t,n_t)$, where $n_t = |y_t|$ and $\sigma_t$ is a distribution on topics $\boldsymbol{\beta}$
    \end{enumerate}
\end{enumerate}
\end{algorithm}

\subsection{A mixed membership RNN topic model}
\label{sec.topic_model}

We extend the basic MM-RNN idea to address the topic modeling problem. Topic models capture semantic meaning through a mixture of $K$ topics $\boldsymbol\beta = \{\beta_1,\dots,\beta_K\}$, being probability distributions on a vocabulary of size $V$. Each document is a set of words generated using a $K$-dimensional mixing weight vector on these topics, $\sigma^{(d)}$ for document $d$. A document $y^{(d)}$ consists of $n_d$ words, where for each word instance a topic index is chosen according to $\sigma^{(d)}$ and the word value is then chosen by drawing from the distribution in $\boldsymbol{\beta}$ with that index. The topics learned are semantically meaningful, and topic models are powerful in that they can be used for far more than text data.

The canonical topic model for sequential data is the dynamic topic model (DTM) \cite{blei2006dynamic}. There, the topics vary in time, while each document generates its own $\sigma^{(d)}$ independently and uses the snapshot of topics at the moment of its generation. This allows prominent words within a coherent topic (e.g., the ``politics'' topic) to evolve over time. Here we consider a different problem where the topics are fixed in time, and the distributions on topics evolve. For example shopping behavior data consists of products (words) in an order (document), and each customer's sequence of orders can be modeled by a mixed membership model where each order's distribution on a fixed set of topics evolves over time.

We describe our general MM-RNN topic model in the table above. The data-generating distribution in Step 4(c) can be the standard mixture of multinomials used by LDA, or it could be a Poisson matrix factorization, or other distribution on count data. To connect this with previous topic models, we observe that if $\rho \equiv 0$ and each group consists of one ``document,'' then this model reduces to the correlated topic model (CTM) \cite{blei2007correlated}. In this sense the proposed model is one possible version of a dynamic CTM.

\subsection{Discussion on model inference}

We have presented our MM-RNN approach in fairly general terms. In this section we discuss two possible instances that we consider in our experiments and discuss an outline of how we optimized them. We discuss MAP optimization for these models.

In our models, we let $f_{\theta}$, used to construct the hidden state $h$, be a single layer LSTM cell as is standard in PyTorch. Let $y_{d,t}$ be a probability vector or histogram, for example constructed from items purchased in order $t$ by customer $d$. Using zero-mean Gaussian priors on all model variables, we can write one possible objective function as
\begin{equation}
\mathcal{L} = \frac{1}{2a}\|\theta\|^2 + \sum_{d=1}^D \frac{1}{2b}\|\phi_d\|^2 +  \sum_{t=1}^{T_d} \frac{1}{2c}\|y_{d,t} - n_{d,t}B\sigma(v_{d,t}) \|^2~~
\end{equation}
where
$$v_{d,t} \equiv \rho_{d,t} h_{d,t} + (1-\rho_{d,t}) \phi_d,$$

and again, $\rho_{d,t}$ is a deterministic, decreasing function of the time between orders ($\rho_{d,1} = 0$). The value $n_{d,t} = |y_{d,t}|$ achieves appropriate scaling for the prediction.

\begin{algorithm}[t]\vspace{5pt}
\textbf{MM-RNN learning outline:} Initialize RNN parameters $\theta$ and initialize all $\phi_d = 0$. Iterate the following:\vspace{-5pt}
\begin{enumerate}
    \item Update each $\phi_d$ via gradient descent\vspace{-5pt}
    \item Update RNN $\theta$ via automatic differentiation\vspace{-5pt}
    \item (optional) Update ``topic'' matrix via multiplicative update. Otherwise fix $B=I$.
\end{enumerate}
\end{algorithm}

We give a rough outline of what the learning algorithm looks like in the table above. We note here that we take the perspective of nonnegative matrix factorization (NMF) using the L2 penalty when the matrix of ``topics'' $B$ is incorporated. In this case, we are doing maximum likelihood on $B$ and the columns do not need to sum to one, yet are still interpretable. $B$ can be learned using the simple multiplicative update strategy of \cite{lee2001algorithms}.




\section{Experiments}\label{sec.experiments}
In this section, we present experiments on the Instacart 2017 online grocery shopping data set.\footnote{\url{instacart.com/datasets/grocery-shopping-2017}} This data consists of 3.4 million orders made by 206K users. The time interval between orders is number of days (capped at 30 days). Each order consists of a count of the number of each product purchased from 50K products and each product belongs to one of 134 aisles. In our experiments, we consider the basic MM-RNN model at the aggregated aisle level, and the MM-RNN topic model at the product level. We train all models on the orders of all customers except for the last order of each customer, which we hold out for prediction to evaluate performance.

We implement our models in PyTorch using automatic differentiation  \cite{paszke2017automatic} and stochastic gradient descent with a learning rate of 0.01. For our selected RNN, we an LSTM with hidden dimension of 10. When $\rho \equiv 1$, our MM-RNN reverts to this LSTM, which is one of the models we compare with. Experiments are done on a cluster node with two NVIDIA Tesla K80 GPUs and 128 GB memory.

\subsection{Aisle level model}\label{sec.aisleresults}
In our first experiment, we consider the basic MM-RNN model of Section \ref{sec.basic_model} on Instacart data aggregated at the aisle level as defined by this online shopping website (e.g., coffee, milk, cereal, tofu meat alternatives---134 aisles in total). Each order is represented as a normalized histogram giving an empirical distribution of that order across the aisles. We use the softmax function for $\sigma$ to predict this distribution for the next order in the sequence. We focus on the L2 penalty at the data level, but observed similar results using the cross entropy of the softmax to the original histogram. Using the function $\rho(\Delta t) = (t_0 + \Delta t)^{-\kappa}$, we set $t_0 \in \{1,10\}$ and experiment with various values of $\kappa$. For each experiment, we learned the model by running 20 epochs over the data, where each epoch took approximately 5 minutes. For each setting we ran 50 experiments with random initialization.

\begin{figure}[t!]
    \centering
    \includegraphics[width=0.7\columnwidth]{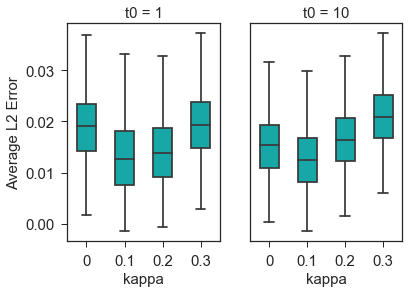}
    \caption{Box plots of MSE as a function of $\kappa$ over multiple runs for $t_0 \in \{1,10\}$. When $\kappa = 0$, the MM-RNN reduces to its base LSTM model. An increase in $\kappa$ indicates that this RNN prediction is being forgotten more quickly as the time between orders increases and the customer-level base distribution is being used instead. As is evident, a combination of sequential/non-sequential modeling gives more accurate predictions.}
    \label{fig:lastorder-pred-on-kappa-aislelevel}
\end{figure}

In Figure \ref{fig:lastorder-pred-on-kappa-aislelevel} we show box plots of average L2 error over the 206K customers' predictions as a function of $\kappa$. As mentioned, when $\kappa = 0$, the MM-RNN reduces to its base LSTM model. An increase in $\kappa$ indicates that this RNN prediction is being forgotten more quickly as the time between orders increases and the customer-level base distribution is being used. We see that performance improves as $\kappa$ increases, followed by a decrease in performance. Clearly for this data a combination of sequential/non-sequential modeling that takes into consideration customer-level effects and the time between orders is appropriate.

\begin{figure}[h!]
	\centering
	\includegraphics[width=0.5\columnwidth]{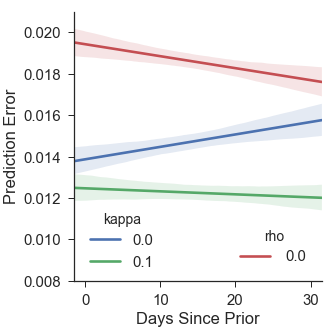}
	\caption{The L2 error between the aisle distribution and predicted distribution for the last order as a function of time passed since the previous order. We use our basic MM-RNN model (with $\kappa = 0.1$) and compare with an LSTM RNN (equivalent to $\kappa = 0$). As is evident, the LSTM decreases in predictive performance as more time passes between observations. When $\rho \equiv 0$, the model reduces to an exchangeable model on orders, giving further support to our belief in the decreasing sequential value as time lag increases.}
	\label{fig:lastorder-pred-days-since}
\end{figure}

In Figure \ref{fig:lastorder-pred-days-since}, we break down these results for $\kappa \in \{0, 0.1\}$ and $t_0 = 1$ using the output of the run closest to the mean of their corresponding box plots in Figure \ref{fig:lastorder-pred-on-kappa-aislelevel}. We also show results for $\rho \equiv 0$, which reduces the MM-RNN to an exchangeable, i.i.d.\ model conditioned on $\phi_d$ for customer $d$. Here, we show the mean and standard deviation of the prediction errors as a function of days between the previous order and the predicted order.

As we expected, the RNN ($\kappa = 0$) makes worse predictions as this time lag increases, likely because it relies completely on previous sequential information that is less useful in this case. The MM-RNN ($\kappa = 0.1$) is able to adapt and focus more on using the base distribution defined by $\phi_d$ for customer $d$. In fact, the performance slightly improves, perhaps indicating that as more time passes the customer runs out of more things and makes and order based on a non-sequential distribution on aisles representing that customer's overall preference. In other words, guessing precisely what a customer needs next is inherently more difficult than guessing what that customer needs ``when the cupboard is empty.'' The RNN does not adapt well here, while our simple modification does. We also observe that when the time lag decreases our model still outperforms the RNN. This may be due to the fact that the learned RNN in the MM-RNN was able to better focus on the meaningful sequential content in the data during inference, while the vanilla RNN considers all parts of the sequence as equally meaningful.

Significantly, when $\rho = 0$ we see the same MM-RNN pattern, only worse since no sequential information is being modeled. As time lag increases, the observations from a customer are more approximately conditionally i.i.d., while when the time lag decreases sequential information is important when considering what does and doesn't need to be purchased. This shows that our approach can meaningfully adapt by blending sequential and non-sequential information in the data.


\subsection{Product level model}
We also experiment at the product level using the MM-RNN topic model discussed in Section \ref{sec.topic_model}. To initialize the non-negative topic matrix $B$, we run stochastic LDA \cite{hoffman2013stochastic} on the individual orders as documents to learn 25 topics and use the means of their respective $q$ distributions as initialization. We use the products as vocabulary, but we aggregate products that were purchased less than 20 total times by their aisle. As a result, $B$ is an approximately $36K \times 25$ matrix with topics on the columns. When we ran the MM-RNN model, we then updated $B$ using the multiplicative update rule of \cite{lee2001algorithms}.

For this experiment we also compare with various imputation strategies described in \cite{pham2016deepcare,lipton2016directly}. We call these three techniques: 1) Impute Mean, which fills in any missing time step with the global mean; 2) Impute Forward, which fills in missing time points with a copy of the most recent observation; 3) Impute Zero, which fills in missing time points with a vector of zeros. We also compare with \cite{che2018recurrent} (GRU-D), a recent method that also uses a continuous-time weighting strategy to account for different time lags. However, this approach does not take a mixed membership perspective by learning group-level parameters, and the weighting strategy is within the RNN itself, rather than outside of the RNN as in our MM-RNN model. Finally, we compare with the LSTM-RNN ($\kappa = 0$).

\def\arraystretch{1.25}
\begin{table}[t!]
	\caption{MSE of predictions at the product level using the MM-RNN topic model. These values correspond to the MSE of the histogram of items in each order divided by the total number of products. As is evident, smoothly modeling the time lag using the MM-RNN can significantly outperform various missing data imputation strategies. It also can improve upon GRU-D, which also models the time lag (in a very different way), but learns no user-specific component.}\vspace{10pt}
	\centering
	\begin{tabular}{|l|c|} \hline
   		Method & Mean Estimation Error   \\ \hline \hline
    	Impute Mean & 0.0877 \\ \hline
    	Impute Forward & 0.0539  \\ \hline
    	Impute Zero & 0.0898 \\ \hline
    	GRU-D & 0.0681 \\ \hline
    	LSTM-RNN & 0.0229 \\ \hline
     	MM-RNN & {\bf{0.0153}}  \\ \hline
	\end{tabular}\label{tab:mse}
\end{table}

We show these results in Table \ref{tab:mse}. As is clear, all imputation methods significantly hurt performance by creating unhelpful sequential information for the RNN that do not help the learning or predictions. While GRU-D has performance comparable with the Impute methods, we note that that RNN architecture does not do any group-level modeling, meaning every sequence of orders is treated as being i.i.d.; this indicates the advantage of a mixed membership approach for this type of problem. The MM-RNN also improves over the vanilla RNN with LSTM, which simply ignores the time stamps of the sequences.

\begin{figure}[t!]
    \centering
    \includegraphics[width=0.5\columnwidth]{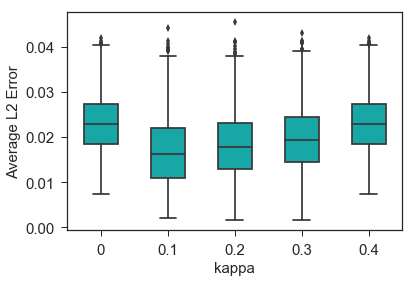}
    \caption{Box plots of MSE normalized by the number of products as a function of $\kappa$ over multiple runs for $t_0 = 1$. These results correspond to the MM-RNN topic model. When $\kappa = 0$, the MM-RNN topic model reduces to its base LSTM model. An increase in $\kappa$ indicates that this RNN prediction is being forgotten more quickly as the time between orders increases and the customer-level base distribution is being used instead. For large values of $\kappa$, the MM-RNN is closely related to the correlated topic model of \cite{blei2007correlated}.}
    \label{fig:lastorder-pred-on-kappa-topiclevel}
\end{figure}

Figure \ref{fig:lastorder-pred-on-kappa-topiclevel} shows the box plots of 50 experiments with random initializations for multiple values of $\kappa$ and $t_0 = 1$. These values are normalized to be the MSE averaged over the 36K dimensions of all 206K predictions. The conclusions for this MM-RNN approach to the dynamic topic model is the same as in Section \ref{sec.aisleresults}: At $\kappa=0$ the model reduces to an LSTM-RNN. We see a clear improvement as $\kappa$ increases, followed by a decline.

\begin{figure}[t!]
    \centering
    \includegraphics[width=0.5\columnwidth]{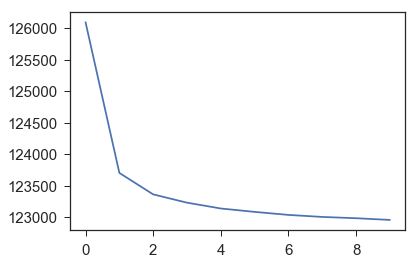}
    \caption{The training objective as a function of the first 10 epochs indicating stable convergence.}
    \label{fig:crossentrop-obj}
\end{figure}

In Figure \ref{fig:crossentrop-obj} we show the objective function for the MM-RNN topic model with $\kappa = 0.1$ as a function of epoch. (Each epoch made 100 updates to the model.) Since each epoch took roughly 5 minutes, this plot represents about 1 hour of computation on a GPU.

\section{Conclusion}
We have presented a mixed membership recurrent neural network (MM-RNN) approach for modeling multiple sequences. The model allows for each sequence to have its own local effects and also adapts to varying time spans between contiguous observations. This allows for a continuous-time handling of what might be inappropriately treated as missing data and imputed. The model is motivated by the observation that, in many sequential data sets the sequential information is not of the same value across the sequence. As more time passes between observations, the distribution on the next observation may be better modeled as independent from some initial distribution with a parameter that depends on the data source (e.g., customer or speaker), similar to the generation of multiple sequences by an HMM.

To this end, we make a simple modification to the RNN architecture. We generate a unique base vector for each group and use a weighted combination of this base vector with the RNN hidden state to make predictions. The weight emphasizes the RNN in the part of  the sequences that is densely sampled, and emphasizes the group-specific i.i.d.\ model when two consecutive observations are spread far apart in time.

We demonstrated on the Instacart online shopping data set that this combination of sequential/non-sequential modeling can allow for the RNN to focus on learning to make better predictions when sequential information is meaningful, and to defer to the base model when much time has passed in a smooth transition. As alluded to previously, we anticipate that this approach can also be usefully applied to other problems, such as ambient speech models.
 
\bibliographystyle{apalike}
\bibliography{reference}

\begin{thebibliography}{}

\bibitem[Airoldi et~al., 2014]{airoldi2014handbook}
Airoldi, E.~M., Blei, D., Erosheva, E.~A., and Fienberg, S. E.~E. (2014).
\newblock {\em Handbook of mixed membership models and their applications}.
\newblock CRC Press.

\bibitem[Bengio and Gingras, 1996]{bengio1996recurrent}
Bengio, Y. and Gingras, F. (1996).
\newblock Recurrent neural networks for missing or asynchronous data.
\newblock In {\em Advances in neural information processing systems}, pages
  395--401.

\bibitem[Blei and Lafferty, 2006]{blei2006dynamic}
Blei, D.~M. and Lafferty, J.~D. (2006).
\newblock Dynamic topic models.
\newblock In {\em Proceedings of the 23rd international conference on Machine
  learning}, pages 113--120. ACM.

\bibitem[Blei et~al., 2007]{blei2007correlated}
Blei, D.~M., Lafferty, J.~D., et~al. (2007).
\newblock A correlated topic model of science.
\newblock {\em The Annals of Applied Statistics}, 1(1):17--35.

\bibitem[Blei et~al., 2003]{blei2003latent}
Blei, D.~M., Ng, A.~Y., and Jordan, M.~I. (2003).
\newblock Latent dirichlet allocation.
\newblock {\em Journal of machine Learning research}, 3(Jan):993--1022.

\bibitem[Che et~al., 2018]{che2018recurrent}
Che, Z., Purushotham, S., Cho, K., Sontag, D., and Liu, Y. (2018).
\newblock Recurrent neural networks for multivariate time series with missing
  values.
\newblock {\em Scientific reports}, 8(1):6085.

\bibitem[Cho et~al., 2014]{cho2014learning}
Cho, K., Van~Merri{\"e}nboer, B., Gulcehre, C., Bahdanau, D., Bougares, F.,
  Schwenk, H., and Bengio, Y. (2014).
\newblock Learning phrase representations using rnn encoder-decoder for
  statistical machine translation.
\newblock {\em arXiv preprint arXiv:1406.1078}.

\bibitem[Choi et~al., 2016]{choi2016doctor}
Choi, E., Bahadori, M.~T., Schuetz, A., Stewart, W.~F., and Sun, J. (2016).
\newblock Doctor ai: Predicting clinical events via recurrent neural networks.
\newblock In {\em Machine Learning for Healthcare Conference}, pages 301--318.

\bibitem[Dieng et~al., 2016]{dieng2016topicrnn}
Dieng, A.~B., Wang, C., Gao, J., and Paisley, J. (2016).
\newblock {TopicRNN}: A recurrent neural network with long-range semantic
  dependency.
\newblock In {\em International Conference on Learning Representations}.

\bibitem[Elman, 1990]{elman1990finding}
Elman, J.~L. (1990).
\newblock Finding structure in time.
\newblock {\em Cognitive science}, 14(2):179--211.

\bibitem[Hochreiter and Schmidhuber, 1997]{hochreiter1997long}
Hochreiter, S. and Schmidhuber, J. (1997).
\newblock Long short-term memory.
\newblock {\em Neural computation}, 9(8):1735--1780.

\bibitem[Hoffman et~al., 2013]{hoffman2013stochastic}
Hoffman, M.~D., Blei, D.~M., Wang, C., and Paisley, J. (2013).
\newblock Stochastic variational inference.
\newblock {\em The Journal of Machine Learning Research}, 14(1):1303--1347.

\bibitem[Lee and Seung, 2001]{lee2001algorithms}
Lee, D.~D. and Seung, H.~S. (2001).
\newblock Algorithms for non-negative matrix factorization.
\newblock In {\em Advances in neural information processing systems}, pages
  556--562.

\bibitem[Lipton et~al., 2016]{lipton2016directly}
Lipton, Z.~C., Kale, D., and Wetzel, R. (2016).
\newblock Directly modeling missing data in sequences with rnns: Improved
  classification of clinical time series.
\newblock In {\em Machine Learning for Healthcare Conference}, pages 253--270.

\bibitem[Parveen and Green, 2002]{parveen2002speech}
Parveen, S. and Green, P. (2002).
\newblock Speech recognition with missing data using recurrent neural nets.
\newblock In {\em Advances in Neural Information Processing Systems}, pages
  1189--1195.

\bibitem[Paszke et~al., 2017]{paszke2017automatic}
Paszke, A., Gross, S., Chintala, S., Chanan, G., Yang, E., DeVito, Z., Lin, Z.,
  Desmaison, A., Antiga, L., and Lerer, A. (2017).
\newblock Automatic differentiation in pytorch.
\newblock In {\em NIPS-W}.

\bibitem[Pham et~al., 2016]{pham2016deepcare}
Pham, T., Tran, T., Phung, D., and Venkatesh, S. (2016).
\newblock Deepcare: A deep dynamic memory model for predictive medicine.
\newblock In {\em Pacific-Asia Conference on Knowledge Discovery and Data
  Mining}, pages 30--41. Springer.

\bibitem[Rumelhart et~al., 1986]{rumelhart1986learning}
Rumelhart, D.~E., Hinton, G.~E., and Williams, R.~J. (1986).
\newblock Learning representations by back-propagating errors.
\newblock {\em Nature}, 323(6088):533.

\bibitem[Tresp and Briegel, 1998]{tresp1998solution}
Tresp, V. and Briegel, T. (1998).
\newblock A solution for missing data in recurrent neural networks with an
  application to blood glucose prediction.
\newblock In {\em Advances in Neural Information Processing Systems}, pages
  971--977.

\end{thebibliography}

\end{document}